%% file: main.tex
\documentclass[10pt,twocolumn,letterpaper]{article}

\usepackage[pagenumbers]{cvpr} 

\usepackage{graphicx}
\usepackage{amsmath}
\usepackage{amssymb}
\usepackage{booktabs}
\usepackage{dblfloatfix} 
\usepackage{enumitem}

\usepackage[section]{placeins} 
\usepackage{caption}
\usepackage{subcaption}
\usepackage{float}

\usepackage[pagebackref,breaklinks,colorlinks]{hyperref}
\usepackage[capitalize]{cleveref}
\crefname{section}{Sec.}{Secs.}
\Crefname{section}{Section}{Sections}
\Crefname{table}{Table}{Tables}
\crefname{table}{Tab.}{Tabs.}

\def\cvprPaperID{34} 
\def\confName{CVPR}
\def\confYear{2022}

\usepackage{tensor}
\usepackage{mathabx}

\newcommand{\T}{\mathrm{T}}
\newcommand{\matr}[1]{\mathbf{#1}}
\renewcommand{\vec}[1]{\mathbf{#1}}

\renewcommand{\tilde}[1]{\ensuremath{\widetilde{#1}}}
\renewcommand{\hat}[1]{\ensuremath{\widehat{#1}}}
\renewcommand{\bar}[1]{\ensuremath{\widebar{#1}}}

\begin{document}


\title{Tensor-based Emotion Editing in the StyleGAN Latent Space}

\author{
René Haas, Stella Graßhof, and Sami S.\@ Brandt\\
IT University of Copenhagen\\
{\tt\small \{renha,stgr,sambr\}@itu.dk}
}


\input{sec/intro}

\input{sec/method}

\input{sec/experiments} 

\clearpage
\newpage

{\small
\bibliographystyle{ieee_fullname}
\bibliography{references}
}

\clearpage
\newpage

\appendix

\end{document}

%% file: sec/intro.tex
\twocolumn[{%
\renewcommand\twocolumn[1][]{#1}%
\maketitle
\begin{center}
    \centering
    \captionsetup{type=figure}
    \includegraphics[trim=0 5 0 0,clip, width=\textwidth]{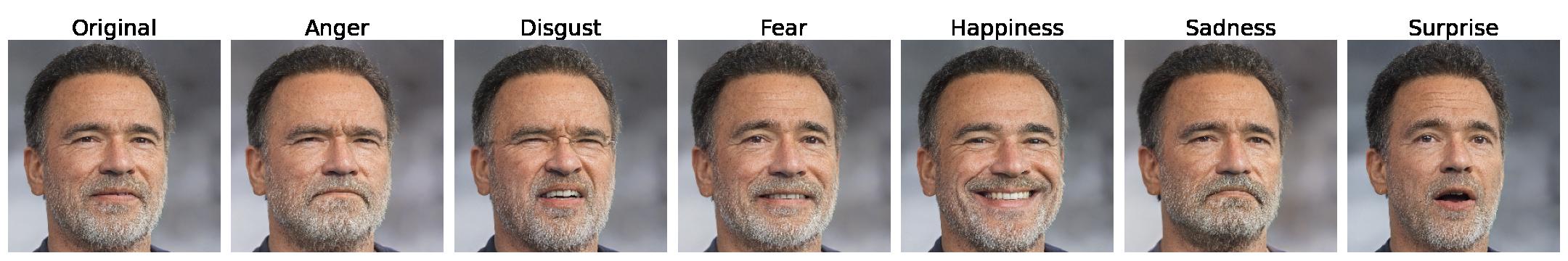}
    \includegraphics[trim=0 25 0 45,clip, width=\textwidth]{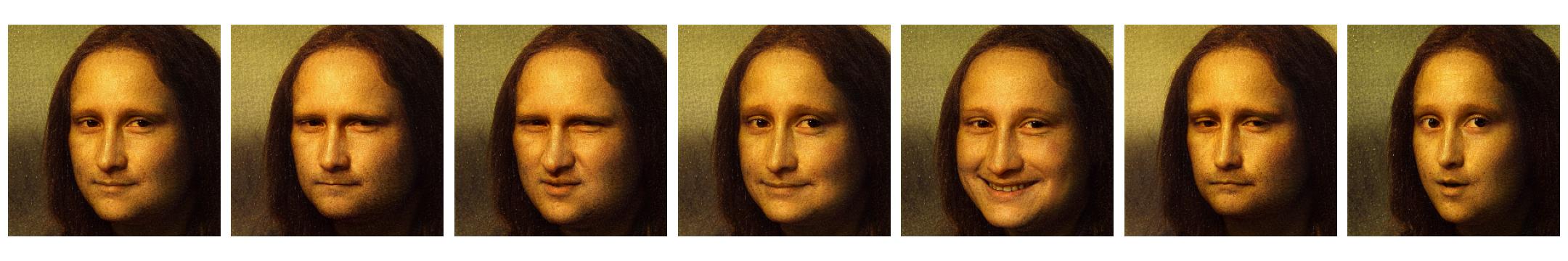}
    \includegraphics[trim=0 45 0 45,clip, width=\textwidth]{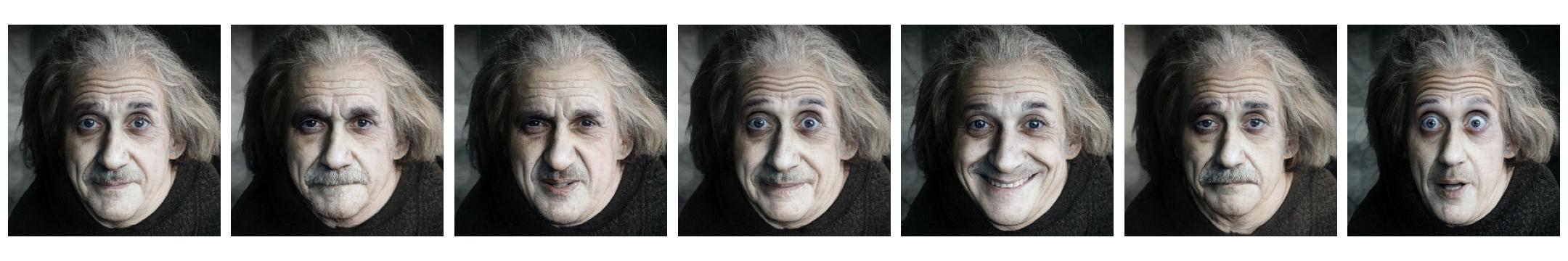}    

    \captionof{figure}{Using our model we can edit StyleGAN latent codes in the direction of the six prototypical emotions.}
   \label{teaser}
\end{center}%
}]

\begin{abstract}
In this paper, we use a tensor model based on the Higher-Order Singular Value Decomposition (HOSVD) to discover semantic directions in Generative Adversarial Networks. 
This is achieved by first embedding a structured facial expression database into the latent space using the e4e encoder. 
Specifically, we discover directions in latent space corresponding to the six prototypical emotions: anger, disgust, fear, happiness, sadness, and surprise, as well as a direction for yaw rotation. 
These latent space directions are employed to change the expression or yaw rotation of real face images. We compare our found directions to similar directions found by two other methods. 
The results show that the visual quality of the resultant edits are on par with State-of-the-Art. 
It can also be concluded that the tensor-based model is well suited for emotion and yaw editing, i.e., that the emotion or yaw rotation of a novel face image can be robustly changed without a significant effect on identity or other attributes in the images.  
\\~ 
\end{abstract}
\vspace{-10pt}

\section{Introduction}
Generative Adversarial Networks (GANs) \cite{Goodfellow2014GAN} have emerged as one of the most promising architectures for image synthesis. 
GANs can produce synthetic images with near-perfect photorealism \cite{Karras2018PGGAN, Karras2019StyleGAN, Brock2019BigGAN, Karras2020StyleGAN2, Karras2020StyleGANada}. 
GANs learn to organize the data they are trained on into a latent space and are, by drawing samples from the latent space, able to synthesize new images which are not contained in the training data but follow the same distribution.
In particular, in the field of face synthesis StyleGAN has set new standards for what is possible \cite{Karras2019StyleGAN,Karras2020StyleGAN2, Karras2020StyleGANada}.

Recent work has explored methods to gain artistic control over the images produced by modern GANs \cite{Shen2020InterfaceganTPAMI,Abdal2020Image2StyleGANpp,Harkonen2020GANSpace,Patashnik2021StyleCLIP,Shen2020SeFa,Tewari2020StyleRig,Wu2020StyleSpace,Abdal2020StyleFlow}. 
In this work, we use a multilinear tensor model to derive latent space directions in StyleGAN2 \cite{Karras2020StyleGAN2} corresponding to the six prototypical emotions: anger, disgust, happiness, fear, sadness, and surprise as well as yaw rotation. With these directions, we are able to edit the emotion of real face images as shown in Fig.~\ref{teaser}.

\paragraph{StyleGAN.}
The StyleGAN generator $G$ is composed of two networks, the \emph{mapping network} $f$ and the \emph{synthesis network} $g$. The mapping network $f$ maps the latent vector $\vec{z}\in \mathcal{Z}$ onto the auxiliary latent space $\mathcal{W}$ while the synthesis network maps a vector $\mathbf{w}\in\mathcal{W}$ to the final output image.
The latent vectors in $\mathcal{Z}$ follow the standard normal distribution $\mathcal{N}(\vec{0},\vec{I})$ while the distribution of the auxiliary latent codes in $\mathcal{W}$ is learned by the mapping network $f$. 
The main benefit of this mapping is that the $\mathcal{W}$ space is more disentangled if compared to the $\mathcal{Z}$ space \cite{Karras2019StyleGAN}.

Every major block corresponding to a resolution of the synthesis network is modulated by two style vectors $w_1,w_2\in \mathbb{R}^{512}$. Thus, for the full 1024 by 1024 generator there are 9 major blocks and the synthesis network takes a total of 18 style vectors as an input. Each set of style vectors has different effects on the synthesized image. 
In detail, the style vectors for the early layers, corresponding to coarse spatial
resolutions, control high-level aspects of the image such as pose and face shape. Style vectors on the middle layers control smaller scale facial features like hair style and if the eyes and mouth are open or closed. The style vectors on the later layers correspond to higher resolutions controls such as the texture and the microstructure of the generated image\cite{Karras2019StyleGAN}. 
In $\mathcal{W}$ space, each of the style vectors are identical. 
However, we can also allow them to be different, in which case the resulting space is denoted as the $\mathcal{W}+$ space. 
The $\mathcal{W}+$ space can be used for style mixing \cite{Karras2019StyleGAN} and GAN inversion \cite{Abdal2019Image2StyleGAN, Zhu2020InDomain}. 
Recently, an additional latent space referred to as \emph{style space} has also been proposed \cite{Wu2020StyleSpace}. 

\paragraph{Semantic Face Editing.}
Several methods have been proposed to enable edits of the images produced by StyleGAN. 
InterFaceGAN \cite{Shen2020Interfacegan, Shen2020InterfaceganTPAMI} uses pre-trained binary classifiers to annotate StyleGAN generated images based on single binary attributes, e.g., young vs. old, male vs. female, glasses vs. no glasses. 
Support vector machines are then trained on the annotated data to discriminate between each attribute in the latent space. The normal vectors of the separating hyperplane define a direction in latent space that changes the corresponding binary attribute.
GANSpace \cite{Harkonen2020GANSpace} finds interpretable directions in an unsupervised fashion with PCA while manual examination of the found directions is required. Directions found with PCA are typically entangled, affecting multiple attributes. It was shown that the degree of entanglement can be reduced by only applying the found directions to a subset of the style vectors. 
It has also been proposed to make the eigenvalue decomposition on the weights of the pre-trained generator to discover meaningful semantic directions in the latent space \cite{Shen2020SeFa}.
Recently, StyleCLIP \cite{Patashnik2021StyleCLIP} demonstrates text driven semantic editing by minimizing CLIP \cite{Radford2021CLIP} loss between a text input and the generated image.
StyleFlow \cite{Abdal2020StyleFlow} proposed editing along non-linear paths using  normalizing flows to better preserve identity. 

Separate from StyleGAN research, different multilinear methods have been widely used to model and analyze faces and expressions \cite{Blanz1999MorphableModel,Ferrari2017Dictionary3DMM, tensorface, grasshof2020Multilinear}. 
Recently there has been some interest in applying these methods to explore the latent space of GANs. For example, StyleRig \cite{Tewari2020StyleRig} proposes edits by minimizing the loss between the image produced by the generated image and an image rendered by a 3D morphable model.
Furthermore, models based on the Higher-Order Singular Value Decomposition (HOSVD) have successfully been used to model faces, their 3D reconstruction, as well as in transferring expressions \cite{Vasilescu2002Tensorface, Vlasic2005FaceTransfer, Brunton2014MultilinearWavelets,Chen2014FaceWarehouse}. 
Recently, it has been suggested \cite{Haas2021tensorGAN} to use such a HOSVD-based tensor model for semantic face editing in StyleGAN. 
Here a facial expression database was projected into the StyleGAN $\mathcal{W+}$ space and relevant semantic subspaces corresponding to identity, expression and yaw rotation were defined using HOSVD-based subspace factorization. 
The model showed limited flexibility for representing arbitrary latent codes and to overcome this a stacked style-separated model was proposed. This extended the tensor model to an ensemble of tensor models, one for each style vector in the StyleGAN $\mathcal{W+}$ space.  
Further, it was shown that in the derived expression subspace, each of the six prototypical emotions formed nearly linear trajectories in agreement with \cite{Grasshof2017apathy}. Although initial results were promising, convincing expression editing using a HOSVD-based model on the StyleGAN latent space was however not yet demonstrated. We propose a solution to this shortcoming, and demonstrate the robustness, and competitiveness of our approach in this work.

\paragraph{Generator Inversion.}
To facilitate editing of real images, the images first need to be projected into the StyleGAN latent space. 
This is also referred to as GAN inversion \cite{Zhu2018GANInversion} and the problem is to find a latent code that, when passed to the generator, produces an image as close as possible to the given target image. 
Typically GAN inversion techniques are either based on training an encoder \cite{Pidhorskyi2020ALAE, Richardson2021pSp,Tov2021e4e, alaluf2021restyle}, which can embed an image into latent space at inference time, or optimization-based techniques \cite{Karras2020StyleGAN2,Abdal2019Image2StyleGAN, yang2019unconstrained, Abdal2020Image2StyleGANpp}. 
In the latter approach, the latent code is found by minimizing a loss function, typically pixel-wise L2 or perceptional image similarity \cite{Zhang2018LPIPS} is used. Hybrid approaches have also been proposed which use a trained encoder to find a good initial condition for subsequent iterative optimization of the latent code \cite{puzer, Zhu2020InDomain}.

Recently, \cite{Roich2021pivotal} shows that novel images can be embedded into $\mathcal{W}$ space with a lower reconstruction error by fine-tuning the pre-trained generator on the target image such that the latent code in $\mathcal{W}$ space yields an image closer to the target. 

Recent work \cite{Tov2021e4e} suggests that there is a trade-off between distortion and editability when selecting which latent space to project a given target image into. 
When projecting out-of-domain images into the StyleGAN latent space picking the extended $\mathcal{W}+$ space leads to a higher quality reconstruction, i.e,  it yields an image closer to the target image. 
However, latent codes in the $\mathcal{W}+$ space are generally less editable than latent codes in $\mathcal{W}$ space.
To find latent codes with the optimal trade-off between distortion and editability a novel training methodology was proposed \cite{Tov2021e4e} which embeds images into $\mathcal{W}+$ space in a way that constrains the latent codes to be as close to $\mathcal{W}$ space as possible. 


\begin{figure}[t]
\centering
\includegraphics[width=0.9\linewidth]{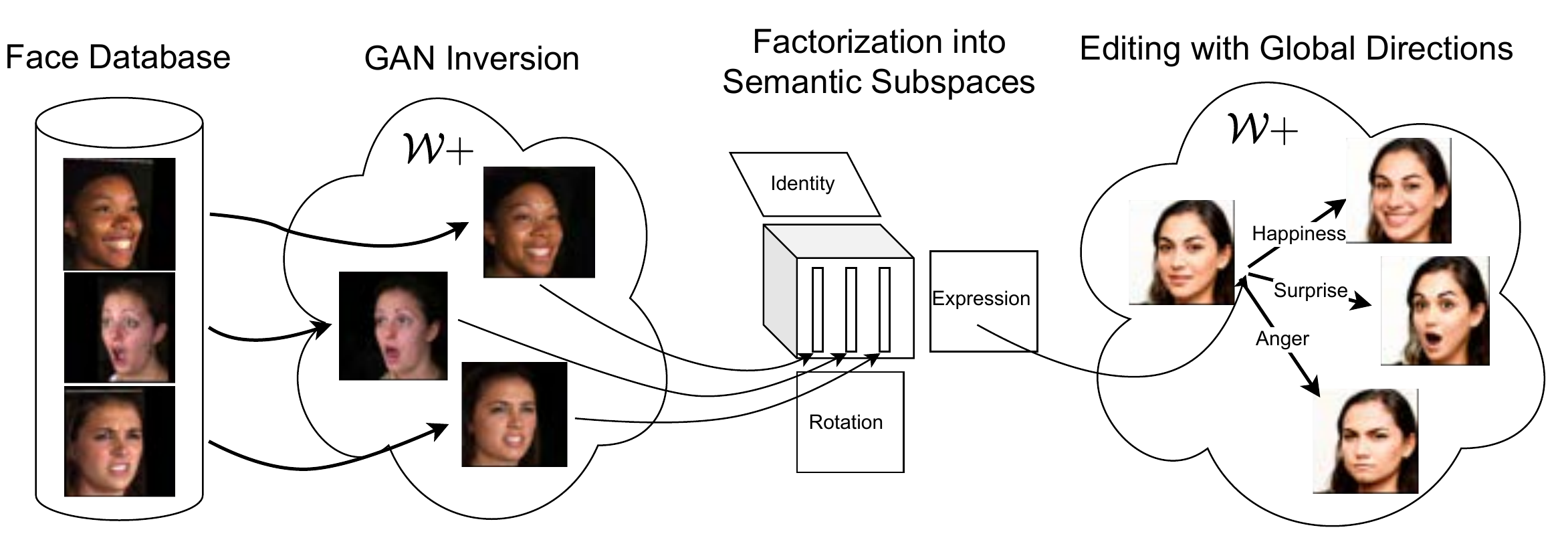}
\caption{Diagram of our method. We first project a facial expression database intro the $\mathcal{W}+$ space of StyleGAN. We then use the HOSVD to factorize the latent representation of the data in order to derive meaningful semantic subspaces. From the subspaces we define a set of global editing directions in $\mathcal{W}+$ corresponding to yaw rotation and each of the six basic emotions.}
\label{fig:method_overview}  
\end{figure}

\paragraph{Contributions.}
Our contributions can be summarized as follows
\begin{itemize}[noitemsep]
\item We show that a HOSVD-based tensor model is able to discover novel semantic directions robustly, corresponding to the six prototypical emotions, in pre-trained GANs.
\item We show that convincing emotion directions can be derived by truncating the expression intensity subspace.
\item We show that, by using the e4e encoder \cite{Tov2021e4e} for projecting real images into the latent space of StyleGAN, it is possible to construct a tensor model which enables stable rotation and expression transfer on real faces. 
\item We show the previously proposed tensor model for the GAN latent space \cite{Haas2021tensorGAN} had an implicit rank-one constraint, which can be relaxed, leading to lower reconstruction error. 
\end{itemize}


%% file: sec/method.tex
\section{Method}

In this section, we describe tensor model formulation \cite{Haas2021tensorGAN} and propose two extensions to it: 
(1) We show how to relax the implicit rank-one constraint of the model by replacing the set of parameter vectors of the model with a single full rank parameter tensor, and 
(2) show how to derive emotion directions in $\mathcal{W}+$ by truncating the expression intensity subspace. 
An overview of our approach is shown in Fig.~\ref{fig:method_overview}.

\subsection{Multilinear Tensor Model}
Given a data set of StyleGAN latent codes in $\mathcal{W+}$ we represent them so that each latent code is equivalent to a vector $\vec{w}\in \mathbb{R}^D$, where $D = 9216$ for the generator producing $1024\times1024$ images. 
Suppose we have latent codes for $P$ different persons, performing $E$ expressions each with $I$ different intensities from $R$ different rotations, then we arrange the data into the 5${^\textrm{th}}$ order tensor $\tensor{T}{} \in\mathbb{R}^{D\times P\times E\times I\times R}$.
We then proceed to calculate the Higher-Order Singular Value Decomposition (HOSVD) on the mean-centered data tensor as
\begin{align}
\tensor{T}{} - \bar{\tensor{T}{}} 
= \tensor{S}{} 
\times_1 \matr{U}_1 \times_2 \matr{U}_2 \times_3 \matr{U}_3\times_4 \matr{U}_4
\times_5 \matr{U}_5, 
\label{HOSVD}
\end{align}
where $\tensor{S}{}$ is the core tensor and $\times_n$ denotes the $n$-mode tensor matrix product. 
The mean tensor is written as $\bar{\tensor{T}{}} = \bar{\vec{w}} \otimes \vec{1}_P\otimes \vec{1}_E\otimes \vec{1}_I \otimes \vec{1}_R$, where $\bar{\vec{w}}$ is the mean latent code from the data set, $\vec{1}_P$ is a vector of ones with dimension $P$, and $\otimes$ denotes the tensor product.
The $\matr{U}_i$ matrices have orthonormal columns, i.e., $\matr{U}_i^\T \matr{U}_i = \matr{I}$ and are constructed from the left singular vectors of the mode-$n$ matrix unfoldings of the mean-centered data tensor.
The columns of $\matr{U}_i$ form the basis for the respective subspace. 
The columns of $\matr{U}_1$ form a basis for the latent space and are identical to the principal components \cite{grasshof2020Multilinear}.
Likewise $\matr{U}_2$, $\matr{U}_3$, $\matr{U}_4$, and $\matr{U}_4$ form the bases for the person identity, expression, intensity and rotation subspaces respectively. 

\paragraph{Parameter Vectors.}
To recover a specific latent code from the tensor model, we select appropriate rows of $\matr{U}_2$, $\matr{U}_3$, $\matr{U}_4$ and $\matr{U}_5$ corresponding to the desired person, expression, expression intensity, and rotation respectively. 
By introducing one-hot vectors $\vec{q}_i^\prime$ which we will refer to as the \emph{canonical} parameters for the tensor model, we get
\begin{align}
\hat{\vec{w}} = \bar{\vec{w}} + 
\tensor{C}{}
\times_2 {\vec{q^\prime}}_2^{\T} \matr{U}_2 
\times_3 {\vec{q^\prime}}_3^{\T} \matr{U}_3
\times_4 {\vec{q^\prime}}_4^{\T} \matr{U}_4
\times_5 {\vec{q^\prime}}_5^{\T} \matr{U}_5,
\label{eq:canonical_model}
\end{align}
where $\tensor{C}{} = \tensor{S}{}\times_1\matr{U}_1$. 
This formulation is analogous to the one proposed in \cite{Grasshof2017apathy,grasshof2020Multilinear} and subsequently, \cite{Haas2021tensorGAN}.
Now, \eqref{eq:canonical_model} can be further simplified by defining $\vec{q}_i^{\T} = {\vec{q}'}_i^{\T} \matr{U}_i$ which allows us to write
\begin{align}
   \hat{\vec{w}}  = \bar{\vec{w}} +\tensor{C}{}
\times_2 \vec{q}_2^\T
\times_3 \vec{q}_3^\T
\times_4 \vec{q}_4^\T
\times_5 \vec{q}_5^\T,
\label{eq:matrix_model} 
\end{align}
which gives is a more compact representation of the tensor model.

\paragraph{Recovering Subspace Parameters.}
To find the parameters $(\vec{q}_2,\vec{q}_3, \vec{q}_4, \vec{q}_5)$ for a novel latent code $\vec{w}$, with corresponding to the latent code $\hat{\vec{w}}$ which best approximates $\vec{w}$, one could 
minimize the $L_2$ loss,
\begin{align}
\mathcal{L}(\vec{q}_2,\vec{q}_3,\vec{q}_4, \vec{q}_5) = ||\hat{\vec{w}} (\vec{q}_2,\vec{q}_3,\vec{q}_4, \vec{q}_5)-\vec{w}||^2_2.
\label{eq:optim}
\end{align}
Additionally, it has been proposed in \cite{Grasshof2017apathy} to regularize the solution by the Tikhonov regularizer and sum constraint as
\begin{align}
   \mathcal{R}(\vec{q}_2,\vec{q}_3,\vec{q}_4, \vec{q}_5) =  \sum_{i=2}^5 \left[ \lambda_{1,i} ||\vec{q^\prime}_i^\T||^2_2 + \lambda_{2,i}(\vec{q^\prime}_i^\T\vec{1} - 1 )^2 \right ],
\label{regularization}
\end{align}
that yields the regularized minimization problem
\begin{align}
   \min_{\vec{q}_2,\vec{q}_3,\vec{q}_4, \vec{q}_5}
\mathcal{L}(\vec{q}_2,\vec{q}_3,\vec{q}_4, \vec{q}_5) + 
\mathcal{R}(\vec{q}_2,\vec{q}_3,\vec{q}_4, \vec{q}_5).
\end{align}
This regularization is important for finding a stable parameter vector representations and thereby enables expression editing for latent codes corresponding to novel images, as will be seen below.  

\paragraph{Relaxing the Rank-One Constraint.}
In the tensor model \eqref{eq:matrix_model}, each latent code is entirely determined by four parameter vectors $\vec{q}_2$, $\vec{q}_3$, $\vec{q}_4$ and $\vec{q}_5$ corresponding to identity, expression, expression intensity and rotation, respectively. 
Using component notation and the Einstein summation convention we rewrite \eqref{eq:matrix_model} as 
\begin{align}
   \hat{w}_i  = 
   \bar{w}_i  + 
   C_{ijklm}
   q_j^{(2)}q_k^{(3)}q_l^{(4)} 
   q_m^{(5)},
   \label{eq:componentmodel}
\end{align}
where $Q_{jklm}= q_j^{(2)}q_k^{(3)}q_l^{(4)}q_m^{(5)}$ is a rank-one tensor.

Now, we propose to relax this implicit rank-one constraint and instead allow the tensor $Q_{jkl}$ to be full rank that leads to the problem
\begin{align}
   \min_{Q} ||\hat{\vec{w}}(Q)-\vec{w}||^2_2.
\end{align} 
The relaxation increases the number of parameters of the tensor model from $P + E + I + R$ parameters to $PEIR$ parameters. This results in a more flexible model which yields lower reconstruction errors for novel latent codes.


\subsection{Truncating the Expression Intensity Subspace}
From \eqref{HOSVD}, the expression intensity subspace is truncated to a one-dimensional subspace by selecting the dominant singular vector, i.e., the first column of $\matr{U}_4$ which we denote $\tilde{\vec{u}}_4$. 
The truncated core tensor is then written as 
\begin{align}
\tilde{S} = (\tensor{T}{} - \bar{\tensor{T}{}}) 
\times_1 \matr{U}_1^\T  
\times_2 \matr{U}_2^\T 
\times_3 \matr{U}_3^\T 
\times_4 \tilde{\vec{u}}_4^\T 
\times_5 \matr{U}_5^\T.
\end{align}
Defining $\tilde{C} = \tilde{S} \times_1 \matr{U}_1$ as before, then the model is written similarly to \eqref{eq:canonical_model} and \eqref{eq:matrix_model} as 
\begin{align}
\hat{\vec{w}} 
&= \bar{\vec{w}} + \tilde{C}
\times_2 {\vec{q^\prime}}_2^{\T} \matr{U}_2 
\times_3 {\vec{q^\prime}}_3^{\T} \matr{U}_3
\times_4 {\vec{q^\prime}}_4^{\T} \tilde{\vec{u}}_4
\times_5 {\vec{q^\prime}}_5^{\T} \matr{U}_5,
\label{eq:lin_model}
\end{align}
where the corresponding intensity parameter  ${\vec{q^\prime}}_4^{\T} \tilde{\vec{u}}_4 = q_4$ is a scalar since the expression intensity subspace has been truncated. 
Thus, the expression intensity factors out of the model and we may write 
\begin{align}
\hat{\vec{w}} 
&= \bar{\vec{w}} + 
q_4(\tilde{C} 
\times_2 \vec{q}_2^\T
\times_3 \vec{q}_3^\T 
\times_5 \vec{q}_5^\T),
\end{align} 
where $q_4$ can now be interpreted as the expression intensity parameter. 
We trivially unfold the singleton dimension of $\tilde{C}$ corresponding to the intensity subspace, i.e., $\tilde{C}_{ijklm}\to \tilde{C}_{ijkm}$ and then write the model as 
\begin{align}
   \hat{w}_i = \bar{w}_i + q^{(4)}\tilde{C}_{ijkm} q_j^{(2)}q_k^{(3)}q_m^{(5)}.
\end{align}

\subsection{Recovering Semantic Directions}

\paragraph{Emotion Directions.}
We define emotion directions in latent space by selecting an appropriate row $\vec{q}_3^{\text{expr}}$ of $\matr{U}_3$ corresponding to the emotion of interest. 
The combined parameter tensor corresponding to an expression direction is then written as    
\begin{align}
\tensor{Q}{}^\text{(expr)} = \bar{\vec{q}}_2\otimes
\vec{q}_3^{\text{expr}} \otimes\bar{\vec{q}}_5,
\end{align}
where $\bar{\vec{q}}_2$ and $\bar{\vec{q}}_5$ is the mean person and rotation parameters respectively.
To change the expression of a given latent code $\vec{w}$, we interpolate linearly in the direction given by the vector $\vec{n}^\text{(expr)}$ with components
\begin{align}
 n_i^\text{(expr)} = \tilde{C}_{ijkm} Q^\text{(expr)}_{jkm},
\end{align}
thus performing an expression edit as
\begin{align}
   \vec{w}^\text{(expr)}_\text{edit} = \vec{w} + q_4\vec{n}^\text{(expr)}.
\label{expression_editing}
\end{align}

\paragraph{Rotation Direction.}
We edit rotations in a similar way. 
First we select the mean person, expression and expression intensity parameters $\bar{\vec{q}}_2$ $\bar{\vec{q}}_3$ and $\bar{q}_4$ and then define the rotation direction parameter $ \vec{q}_5^{(\text{rot})}$ as the difference between the parameters corresponding to the left and right rotations, i.e., the difference between the two rows of $\matr{U}_5$. We write the rotation direction parameter directly as 
\begin{align}
   \vec{q}_5^{(\text{rot})} = 
   \frac{1}{\sqrt{2}}
   \begin{bmatrix}
      1 \\ -1
   \end{bmatrix}^\T
   \matr{U}_5. 
\end{align}
Now the combined rotation direction tensor is written as
\begin{align}
   \tensor{Q}{}^\text{(rot)} = \bar{q}_4(\bar{\vec{q}}_2\otimes
   \bar{\vec{q}}_3 \otimes \vec{q}^{(\text{rot})}_5),
\end{align}
and we can change the rotation of a latent code as 
\begin{align}
   \vec{w}^\text{(rot)}_\text{edit} = \vec{w} + \beta\vec{n}^\text{(rot)} \quad \text{with} \quad n^\text{(rot)}_i  = \tilde{C}_{ijkm} Q^\text{(rot)}_{jkm},
\end{align}
where $\beta$ is the strength of the rotation.

With this formulation, we apply semantic edits directly in $\mathcal{W+}$ without the need for estimating the tensor model parameters beforehand as has otherwise been suggested \cite{Haas2021tensorGAN}.



%


%% file: sec/experiments.tex

\begin{figure*}[tbh]  
   \centering
   \begin{subfigure}[b]{0.32\linewidth}
   \includegraphics[width=\linewidth]{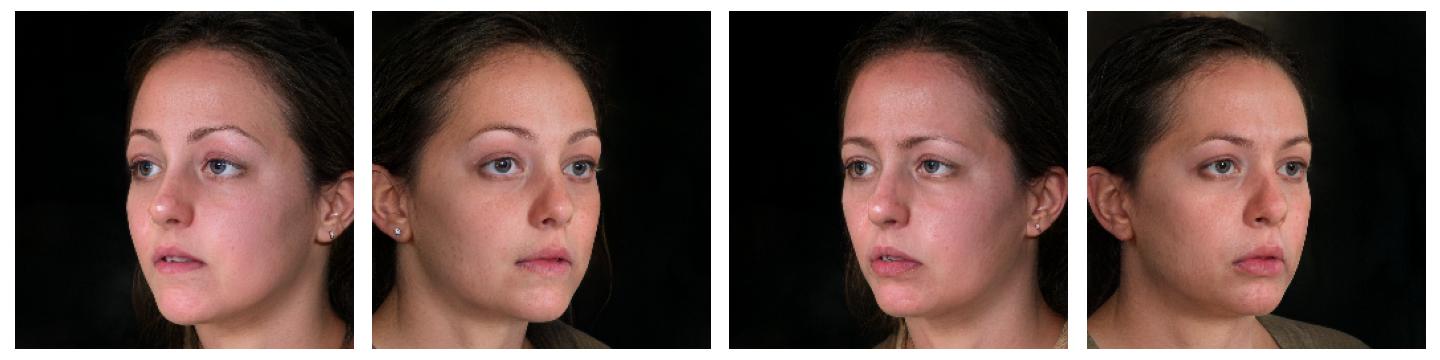}
   \includegraphics[width=\linewidth]{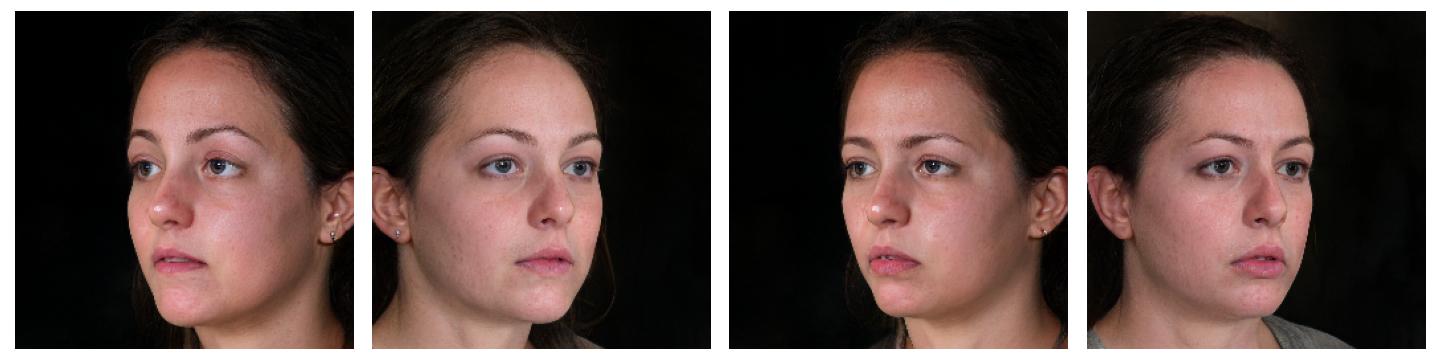}
   \includegraphics[width=\linewidth]{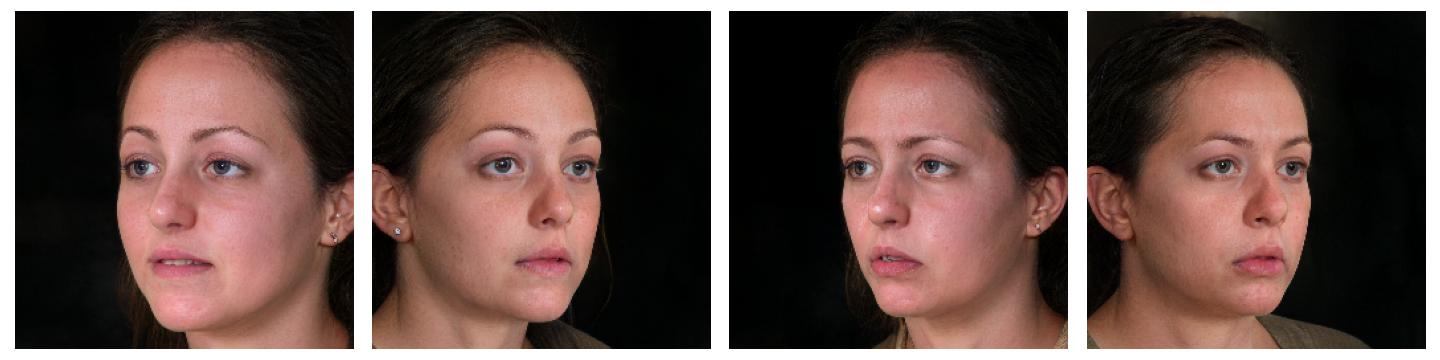}
   \caption{}
   \end{subfigure}
   \begin{subfigure}[b]{0.32\linewidth}
   \includegraphics[width=\linewidth]{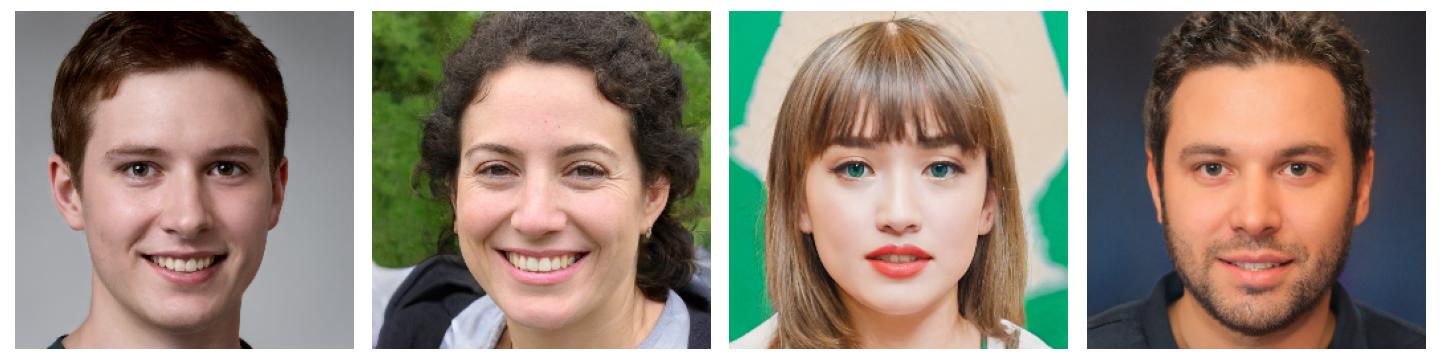}
   \includegraphics[width=\linewidth]{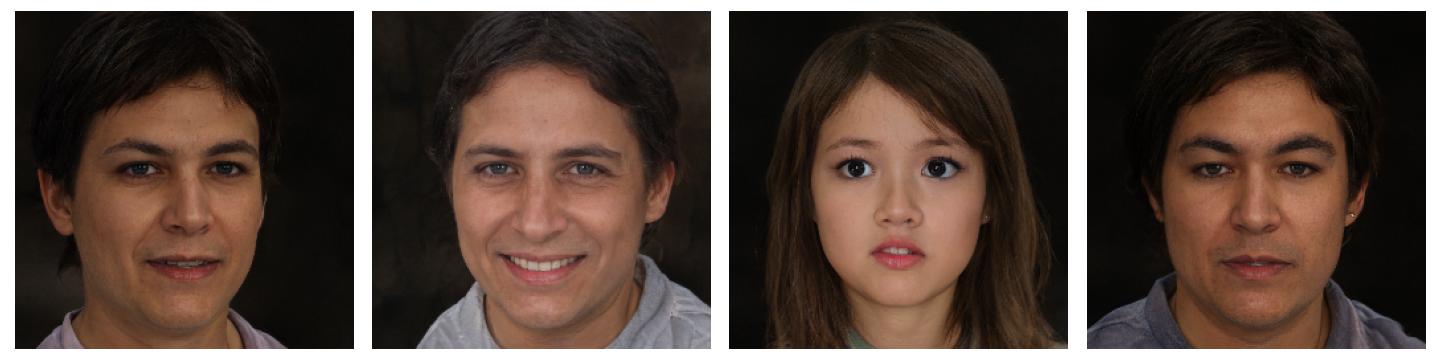}
   \includegraphics[width=\linewidth]{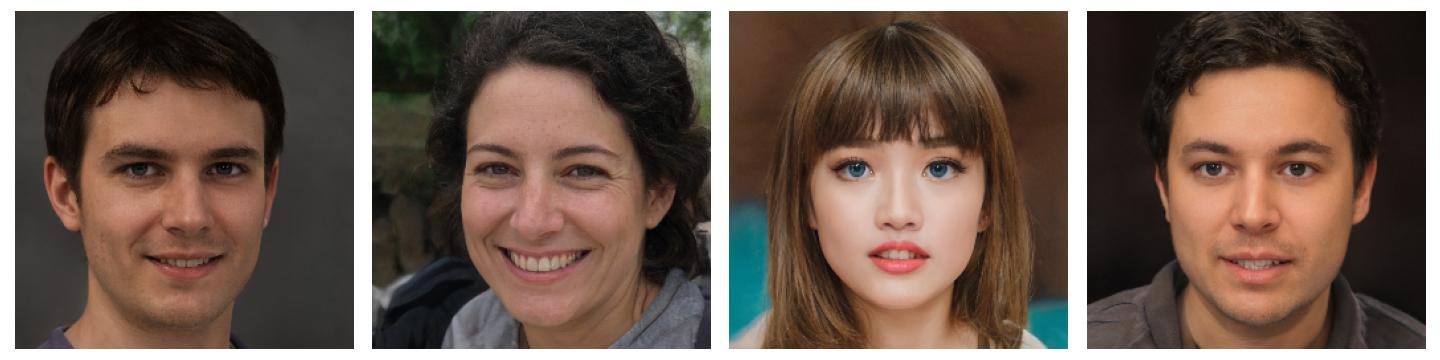}
   \caption{}
   \end{subfigure}
   \begin{subfigure}[b]{0.32\linewidth}
   \includegraphics[width=\linewidth]{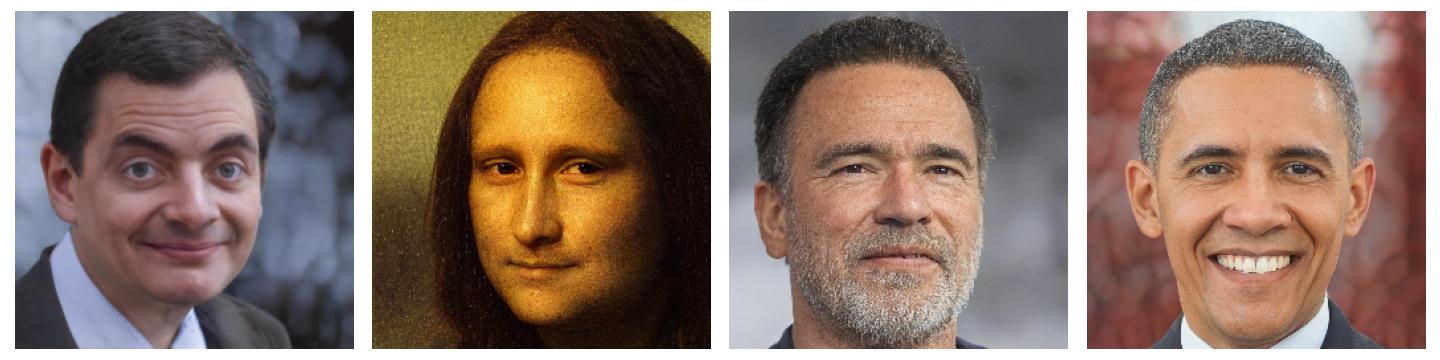}
   \includegraphics[width=\linewidth]{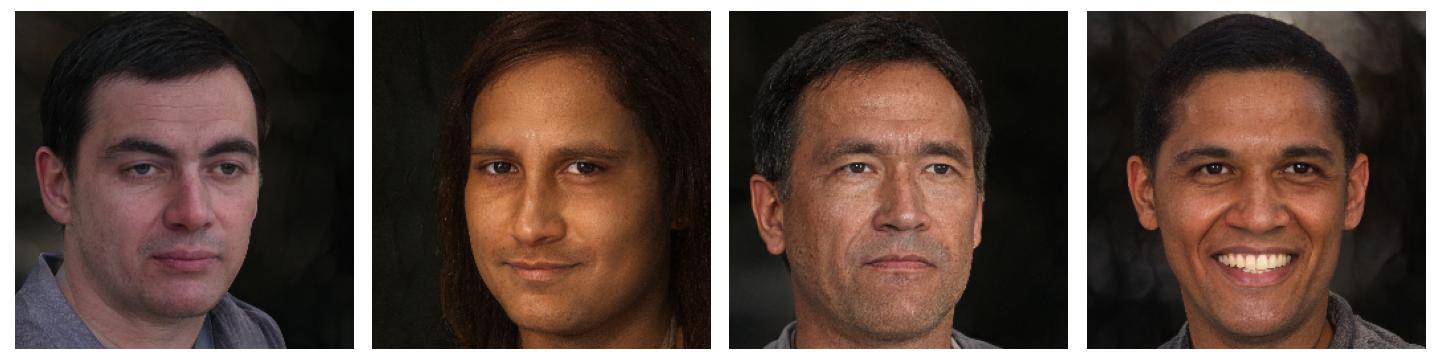}
   \includegraphics[width=\linewidth]{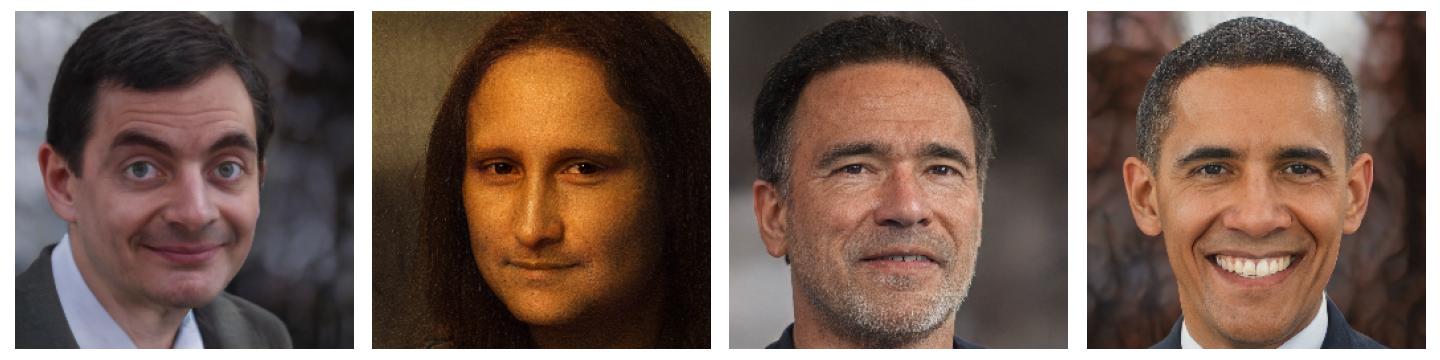}
   \caption{}
   \end{subfigure}
   \caption{Image embeddings. 
   (a) BU-3DFE images, (b) random samples from the generator, and (c) real images.
   The embeddings of the original images are shown in the top row, the parameter vector embeddings in the middle, and the parameter tensor embeddings in the bottom row.}
   \label{recovering_parameters_overview}  
\end{figure*}




\section{Experiments}


Our tensor model was trained with the latent space projection of images from the Binghamton University 3D Facial Expression database (BU-3DFE) \cite{bu3dfe}. 
The BU-3DFE  database contains 2500 3D face scans and corresponding images from two views of 100 persons (56 female and 44  male) with varying ages (18-70 years), and diverse ethnic/racial ancestries. 
Each subject was asked to perform the six basic emotions: anger, disgust, fear, happiness, sadness, and surprise, each with four levels of intensity. 
Additionally, for each participant, a neutral face is provided. 
Hence, for each person, there are 25 facial expressions in total, recorded from two pose directions, left and right, resulting in 5000 face images.   
%
%
Additionally, we used the FEI face database \cite{Thomaz2010Feidatabase} which contains 14 images of each of the 200 individuals, 100 male and 100 female. For each the database contains two frontal images, one with a neutral or non-smiling expression and the other with a smiling facial expression, the rest of the images depicts each individual with a neutral expression from various yaw rotations. 
 
\subsection{Implementation Details}
We use the full resolution, i.e. $1024\times1024$, StyleGAN2 \cite{Karras2020StyleGANada} generator which has been pre-trained on FFHQ \cite{Karras2019StyleGAN}. 
The tensor model was implemented in PyTorch \cite{Paszke2019PyTorch} using tntorch \cite{tntorch} to calculate the HOSVD.
To estimate the tensor model parameters we used gradient descent implemented in PyTorch with the Adam optimizer.
For comparing images we use two different metrics. For perceptual image similarity we use LPIPS \cite{Zhang2018LPIPS} and for identity similarity we uses Arcface \cite{Deng2019ArcFace}.
To measure the pose of the generated images we uses MediaPipe \cite{Lugaresi2019MediaPipe} to extract 2D and 3D landmarks and then proceeded to solve the Perspective-n-point (PnP) \cite{Fischler1981RandomSC} problem which gave us a scalar value for the yaw rotation of a given image. 
We embedded all images into $\mathcal{W}+$ space using the e4e encoder \cite{Tov2021e4e}.




\subsection{Subspace Parameter Recovery}
\begin{figure}[tb]
\centering
\includegraphics[width=\linewidth]{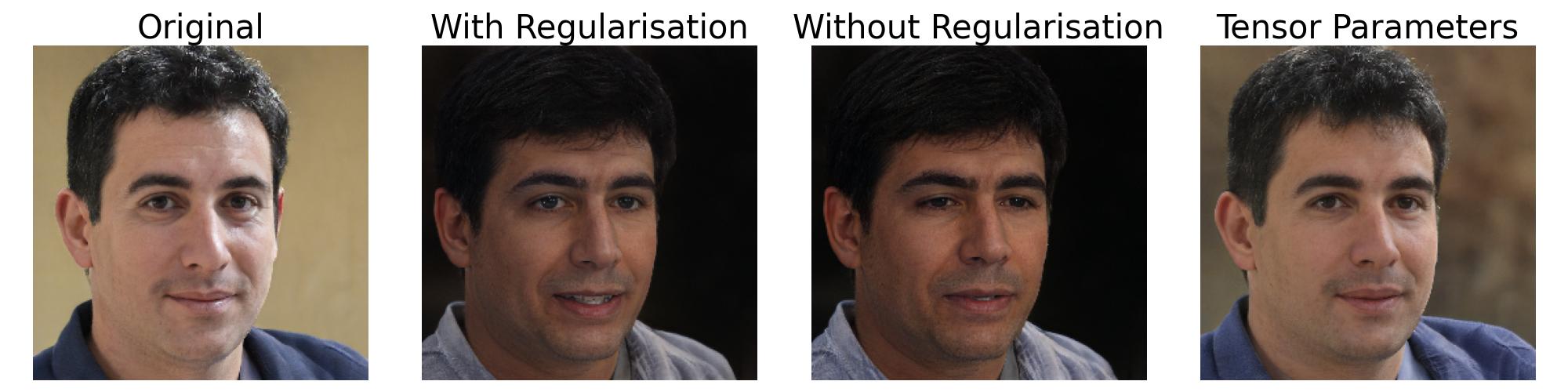}
\includegraphics[trim={0 0 0 2cm},clip,width=\linewidth]{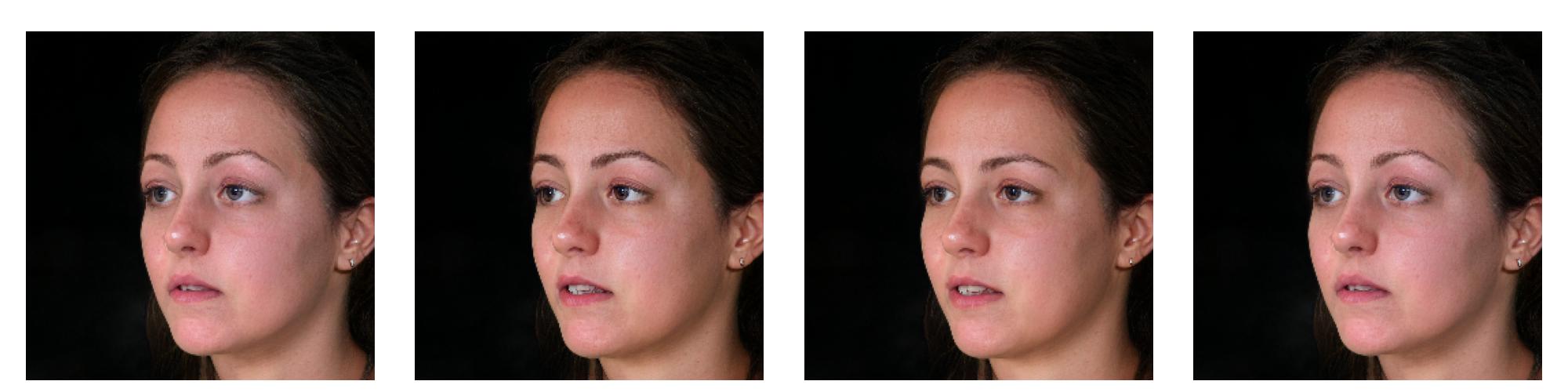}
\caption{Representing a latent code in the tensor model with parameter vectors with and without regularization compared with a representation using a parameter tensor.}
\label{tmembed_compare}
\end{figure}

We computed estimated the tensor model parameters for 3 types of novel latent codes: 1) BU-3DFE latent codes where we left one person out in the calculation of the tensor model, 2) randomly sampled latent codes, and 3) real images projected into latent space. %
Fig.~\ref{recovering_parameters_overview} shows the result of recovering the tensor model parameters for these three types of latent codes when recovering the parameters in vector and tensor form, respectively. It can be seen that using  parameter vectors for the tensor model led to a significant reconstruction loss if compared to using a representation with a parameter tensor, as illustrated in Fig.~\ref{tmembed_compare} and quantified in Tab.~\ref{recon_error}. It seems that the randomly sampled images are slightly harder to reconstruct than the embedded real images.

\begin{figure}[tb]
\centering
\begin{subfigure}{\linewidth}
\includegraphics[width=\linewidth]{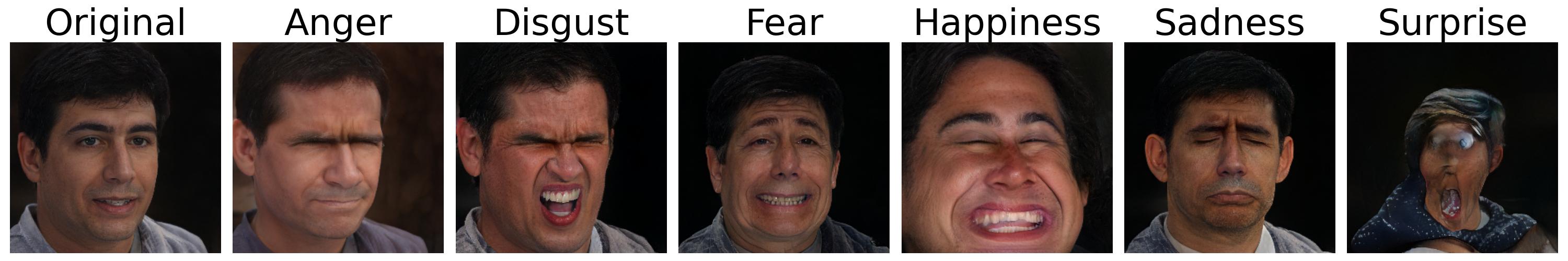}
\includegraphics[trim={0 0 0 2cm},clip, width=\linewidth]{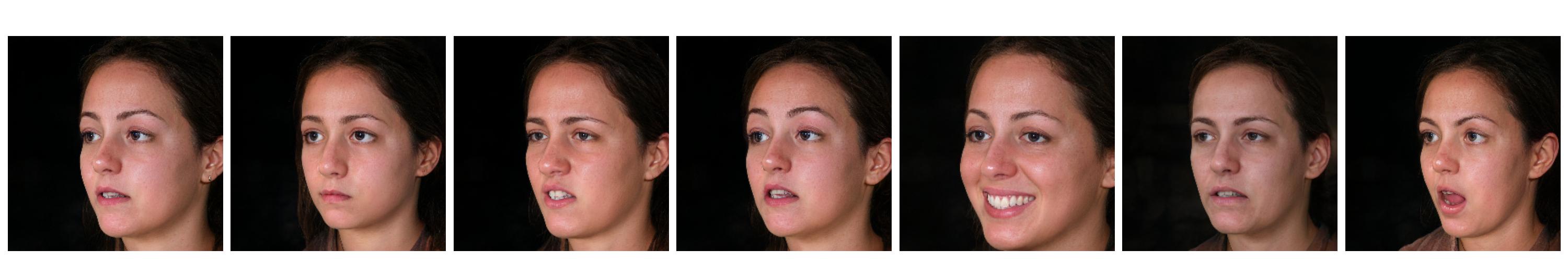}
\caption{Without regularization.}
\end{subfigure}

\begin{subfigure}{\linewidth}
\includegraphics[width=\linewidth]{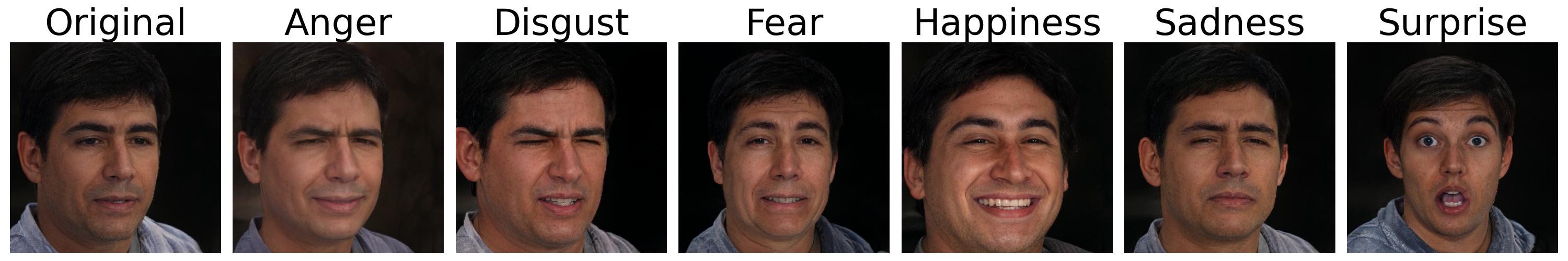}
\includegraphics[trim={0 0 0 2cm},clip,width=\linewidth]{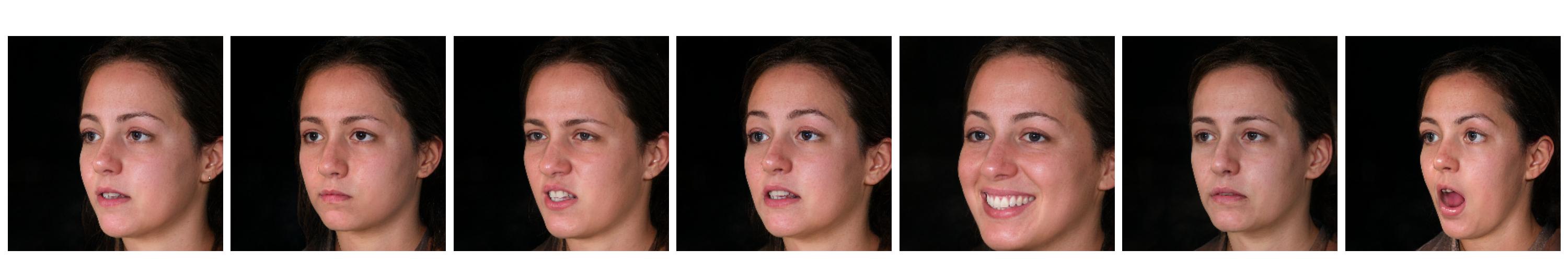}
\caption{With regularization.}
\end{subfigure}
\caption{Visual comparison of the effect of regularization for expression editing using parameter vectors for the tensor model.}
\label{expr_editing}
\end{figure}

For the  representation with parameter vectors, we find that although the proposed regularization \eqref{regularization} leads to a slightly higher reconstruction error, it is important in order to find parameter vectors which are suitable for expression editing. Fig.~\ref{expr_editing} shows that performing expression edits on the regularized parameters leads to less identity change compared to the non-regularized parameters. 
The importance of regularization is more noticeable when we recover the parameters for a randomly generated image if compared to an image contained the in BU-3DFE database.

\begin{table}[tb]
\centering
\caption{Comparison of reconstruction error $||\hat{\vec{w}} - \vec{w}||_2^2$ by representing randomly sampled latent codes and latent codes from the BU-3DFE data set with parameter vector and a parameter tensor respectively.}
\label{recon_error}
\begin{tabular}{||c c c||} 
\hline
& Random Latents & BU-3DFE Latents \\ [0.5ex] 
\hline\hline
Rank one & $(12\pm3)\times 10^2$ &  $(1.7\pm 0.2)\times 10^2$ \\ 
\hline
Full rank &  $(6\pm1)\times 10^2$ & $7 \pm 1$ \\
\hline
\end{tabular}
\end{table}

\begin{figure}[tb]
   \centering
\includegraphics[width=\linewidth]{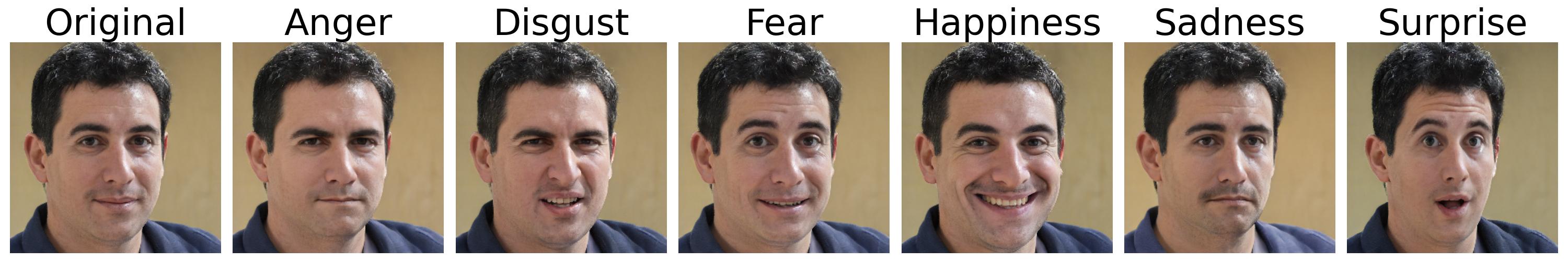}
\includegraphics[trim={0 0 0 2cm},clip,width=\linewidth]{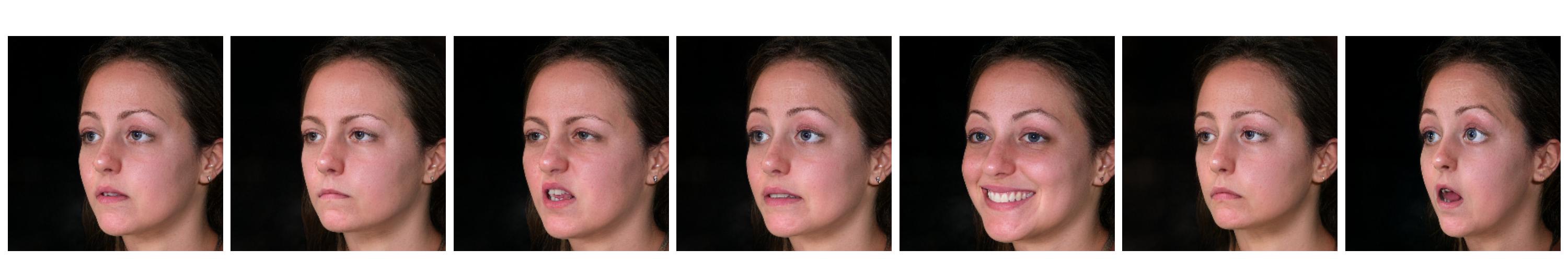}
\caption{Direct edit in the $\mathcal{W+}$ space without prior estimation of the model parameters.}
\label{expr_editing_wspace}
\end{figure}

Moreover, it can be seen that the tensor model is not necessary for expression editing, because we can edit the latent code directly by perturbing in the directions defined by \eqref{expression_editing}, instead of manipulating the estimated parameters of the tensor model.
 The effect of such a direct edit is illustrated in Fig.~\ref{expr_editing_wspace}. The main advantage of performing expression edits in this way, is that we avoid the reconstruction error associated with representing the latent code in terms of the tensor model parameters.

\subsection{Expression Direction Recovery}
\begin{figure}[!b]
\centering
\includegraphics[width=\linewidth]{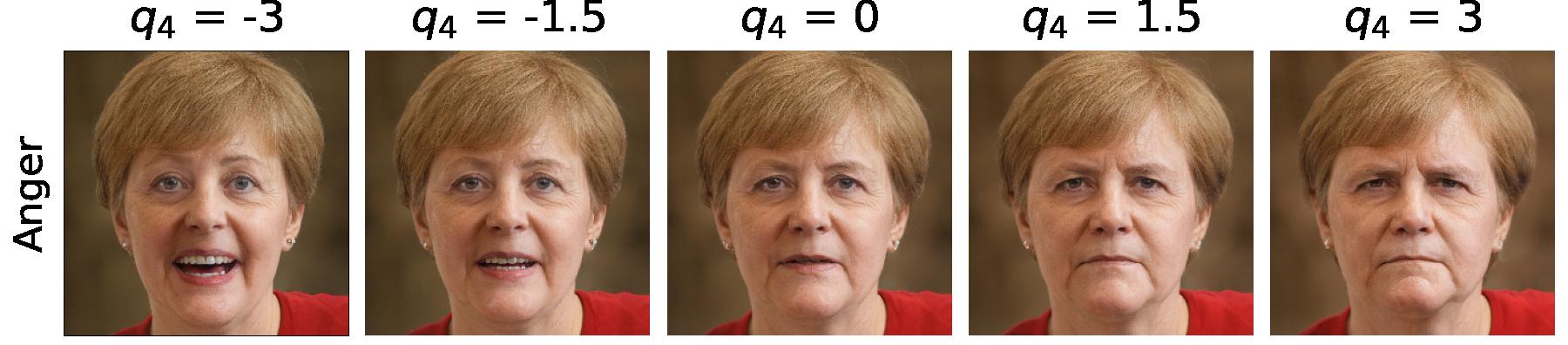}
\includegraphics[width=\linewidth]{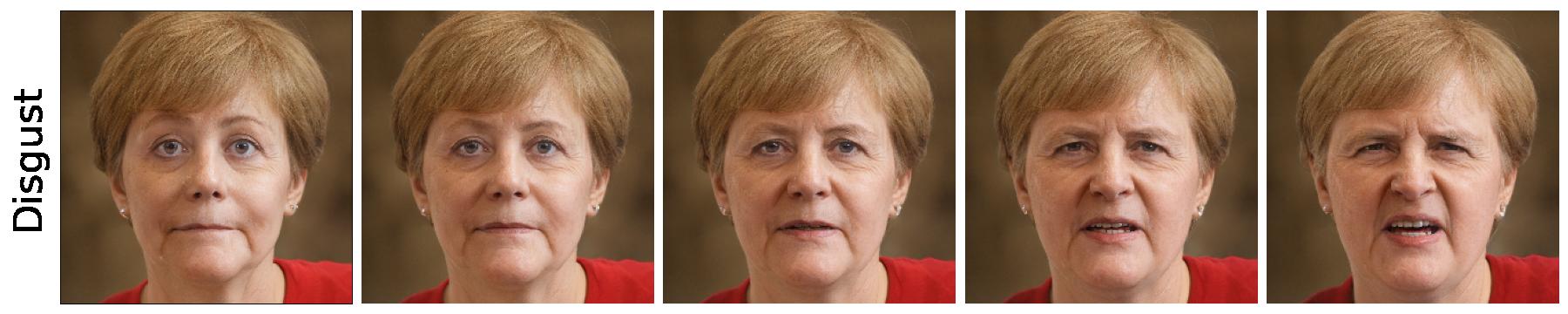}
\includegraphics[width=\linewidth]{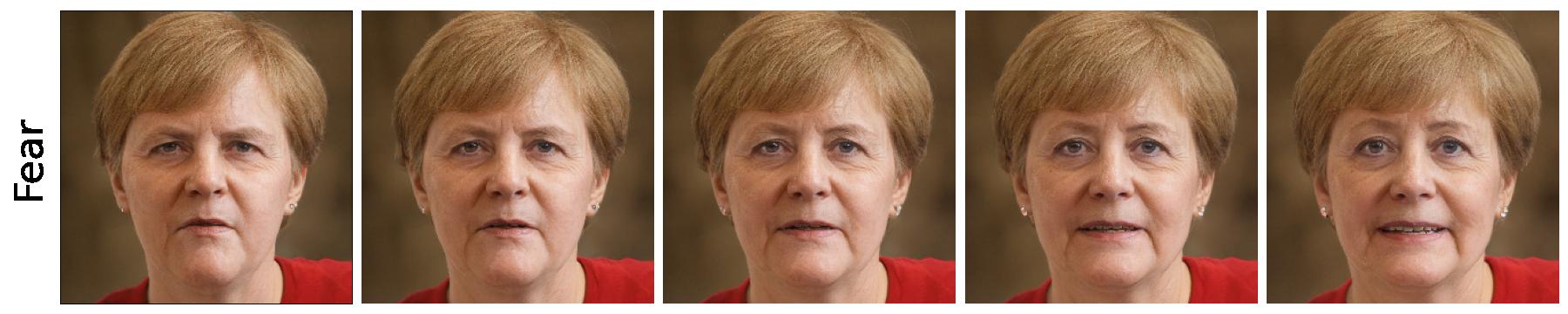}
\includegraphics[width=\linewidth]{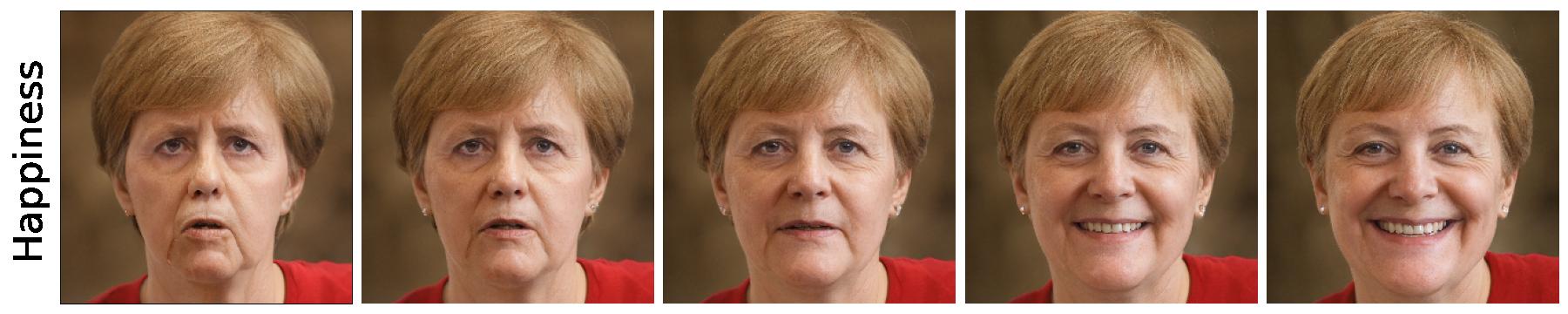}
\includegraphics[width=\linewidth]{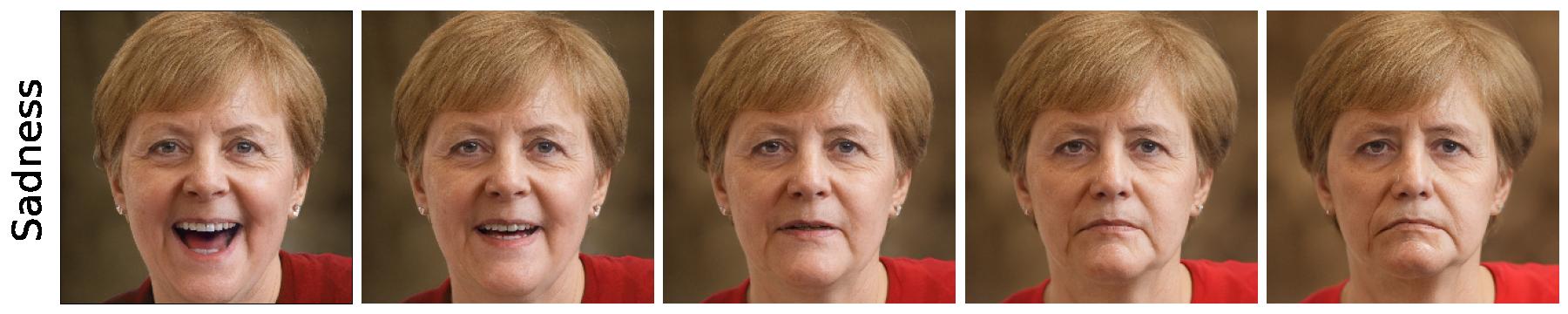}
\includegraphics[width=\linewidth]{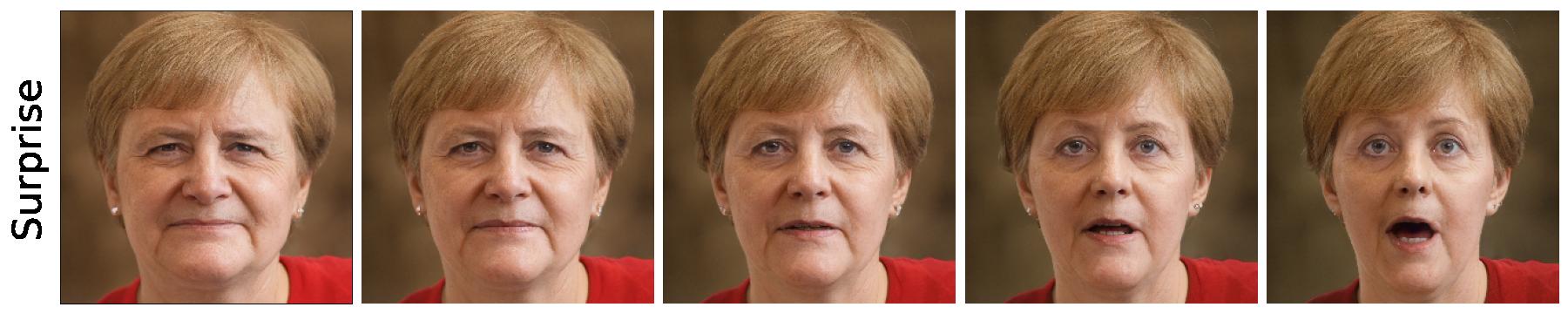}
\caption{Effect of applying the direction corresponding to the six prototypical expressions to a real image. 
The rows show the different expressions determined by $\mathbf{q}_3$ while the strength is modulated by $q_4$, while the rotation parameters $\mathbf{q}_5$ remain unchanged. 
The right column shows edits in the direction of the respective expression while the left column illustrates the subtraction of it. 
}
\label{expression_directions_on_merkel}
\end{figure}

Fig.~\ref{expression_directions_on_merkel} shows the effect of applying the found six latent space directions to the BU-3DFE mean face.
We found that subtracting the sadness direction from the mean face also produces a happy facial expression. 
However, the resulting expression is qualitatively different from adding the happy direction to the mean face. While adding the happy direction results in a wide smile, subtracting the sadness direction results in a smile that is narrower but where the mouth is more open. See the supplementary materials for videos showing the found emotion directions on real face images.  

\subsection{Comparison to Related Work}
\begin{figure}[tb]
\centering
\includegraphics[width=\linewidth]{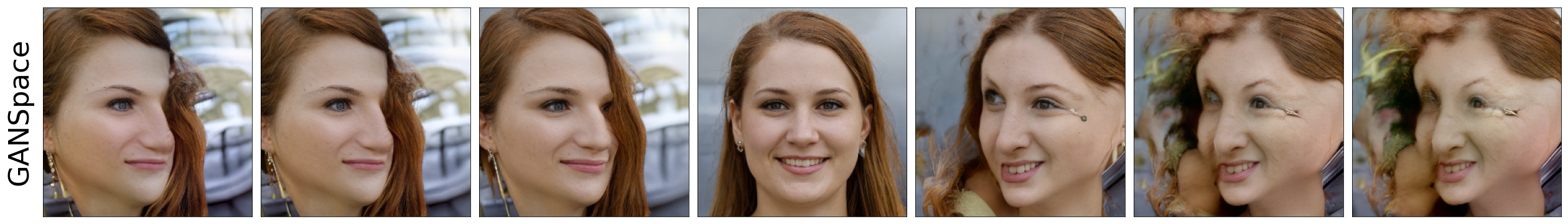}
\includegraphics[width=\linewidth]{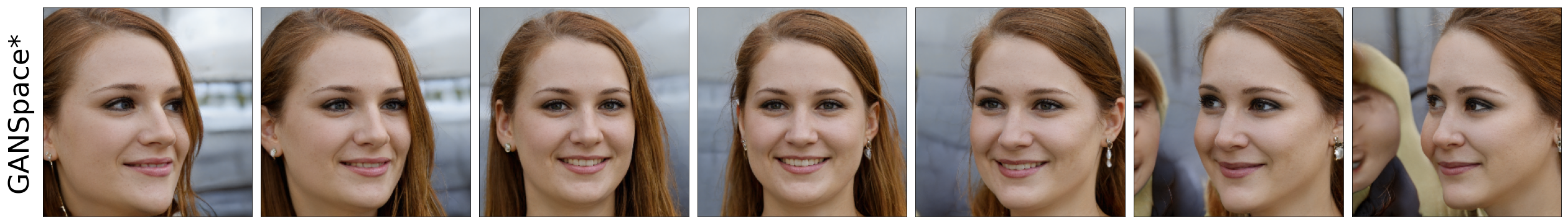}
\includegraphics[width=\linewidth]{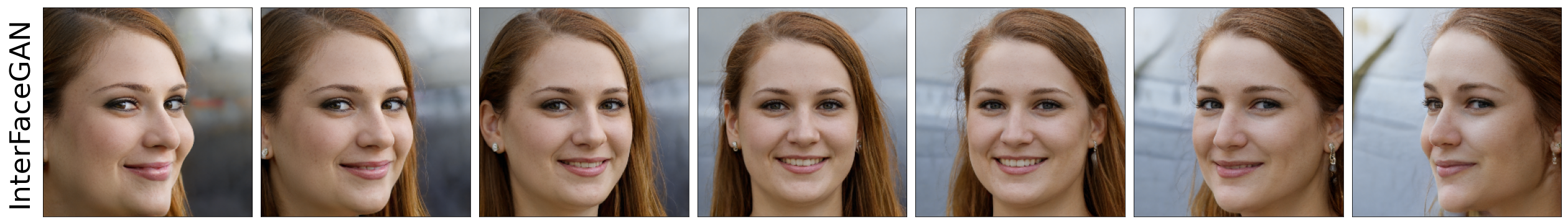}
\includegraphics[width=\linewidth]{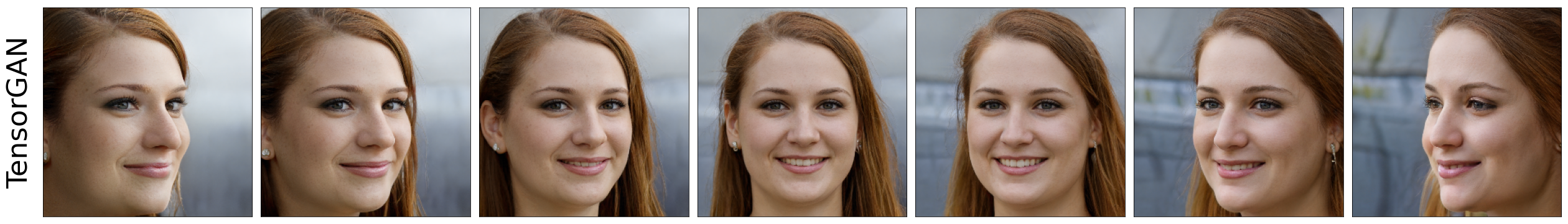}
\caption{Comparison of rotations produced by GANSpace \cite{Harkonen2020GANSpace} (top 2 rows), InterFaceGAN \cite{Shen2020InterfaceganTPAMI} (third row) and our approach (bottom). Here GANSpace* refers to a manipulation where we edit the first five style vectors rather than the first three as described in the main text.}
\label{vis_compare_rotation}
\end{figure}

We compared the rotation and smile directions found by our approach to those previously found by InterFaceGAN \cite{Shen2020InterfaceganTPAMI} and GANSpace \cite{Harkonen2020GANSpace}. 
For InterFaceGAN, we used the PyTorch version of the rotation and smile directions provided by the authors of \cite{Roich2021pivotal} at their GitHub repository\footnote{\url{https://github.com/danielroich/PTI/tree/main/editings/interfacegan_directions}}.
For the rotations, we chose a manipulation strength that resulted in a similar degree of rotation. To perform rotations with GANSpace \cite{Harkonen2020GANSpace}, we initially used the $2^\text{nd}$ principal component applied to the first three style vectors. However, we found that if we only changed the first three style vectors to edit the rotation, the result tends to break down when the editing strength is large, which is demonstrated in the first row in Fig.~\ref{vis_compare_rotation}. 
If we applied the edit to the first five style vectors instead, we generally received better results, see second row in Fig.~\ref{vis_compare_rotation}. 

\begin{figure}[!bt]
\centering
\begin{subfigure}[b]{\linewidth}
\includegraphics[width=\linewidth]{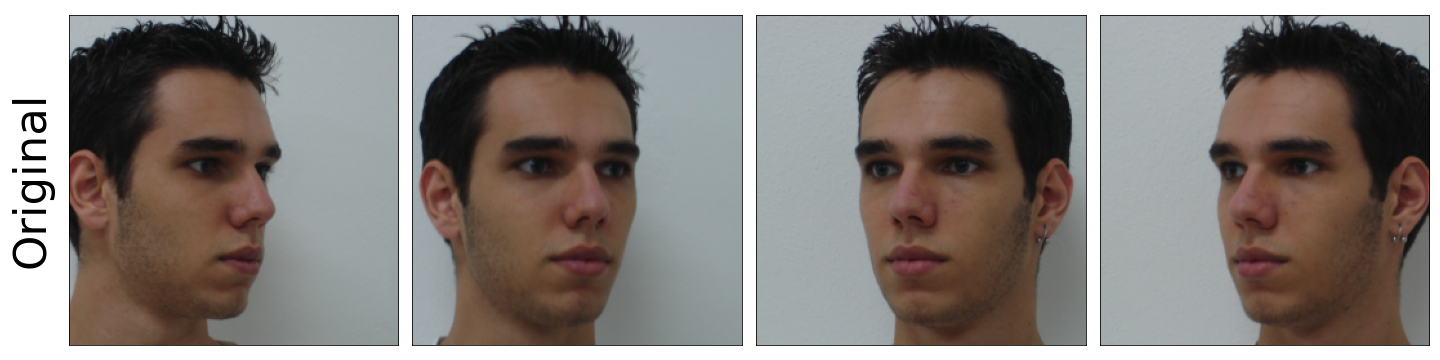}
\end{subfigure}
\begin{subfigure}[b]{\linewidth}
\includegraphics[width=\linewidth]{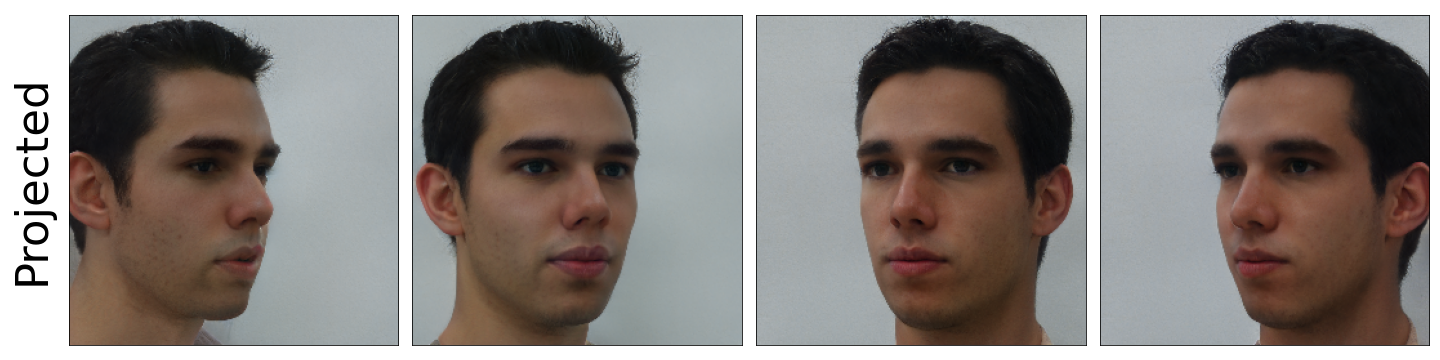}
\end{subfigure}
\begin{subfigure}[b]{\linewidth}
\includegraphics[width=\linewidth]{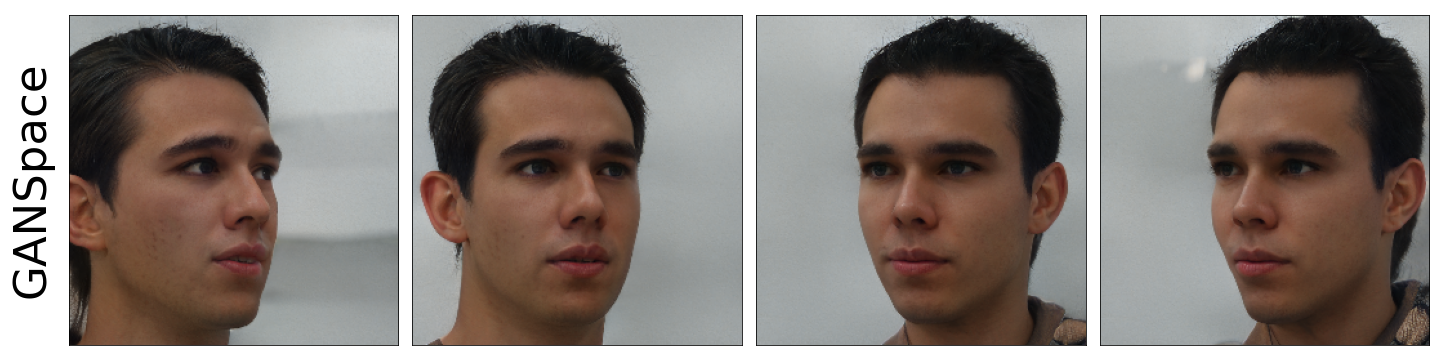}
\end{subfigure}
\begin{subfigure}[b]{\linewidth}
\includegraphics[width=\linewidth]{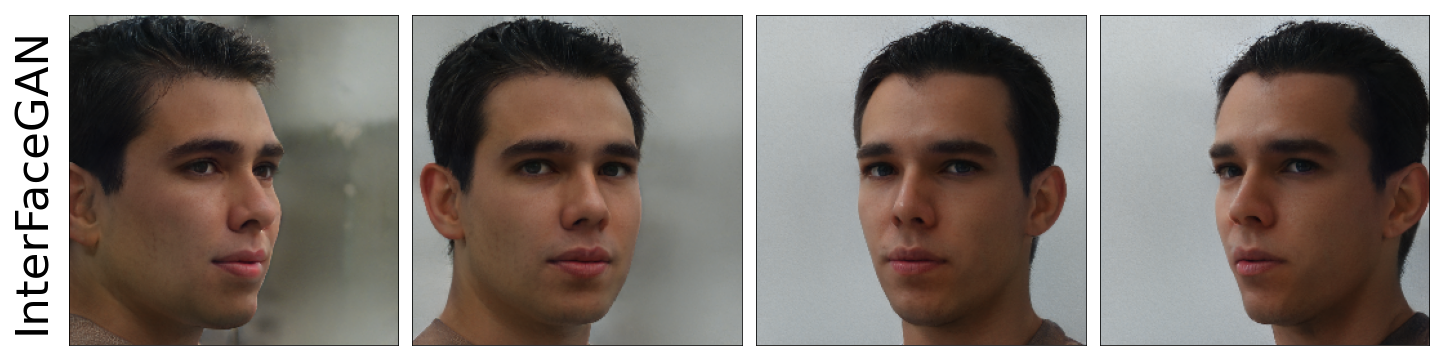}
\end{subfigure}
\begin{subfigure}[b]{\linewidth}
\includegraphics[width=\linewidth]{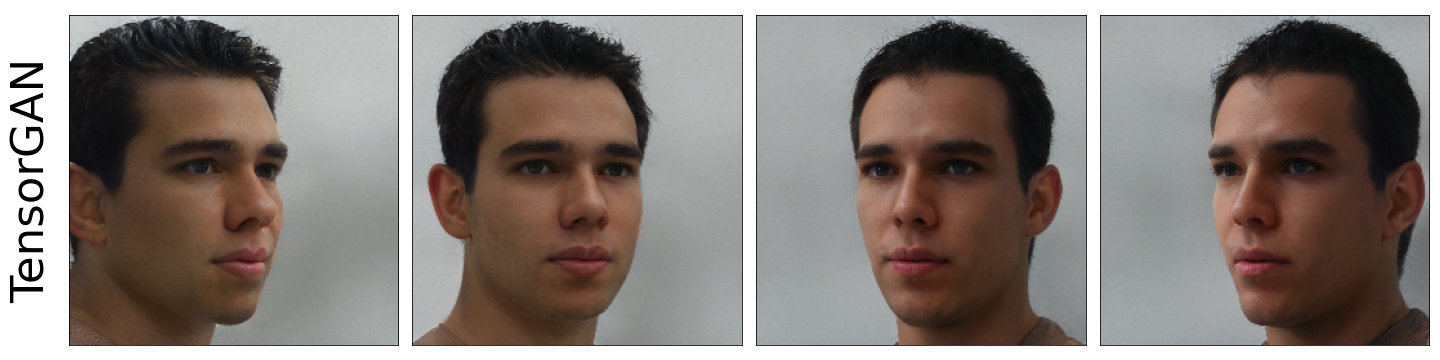}
\end{subfigure}
\caption{Qualitative comparison of the found rotation direction with the equivalent edits from InterFaceGAN \cite{Shen2020InterfaceganTPAMI} and GANSpace \cite{Harkonen2020GANSpace} applied on the FEI face database \cite{Thomaz2010Feidatabase}.}
\label{vis_compare_rotation_fei}
\end{figure}

We visually compared the rotations by GANSpace, InterFaceGAN and our proposed method on images which are randomly sampled from the generator as well as images from the FEI face database \cite{Thomaz2010Feidatabase}.
For the FEI database we used the frontal face images as initial conditions and then applied rotations with GANSpace, InterFaceGAN and our method to approximate the latent codes corresponding to rotated images from the database. 
The results on randomly sampled images are shown in Fig.~\ref{vis_compare_rotation} and on the FEI database in Fig.~\ref{vis_compare_rotation_fei}, respectively.
It can be seen that the quality of the edits are visually on par, except the gaze direction follows the camera in the InterFaceGAN results. 

\subsection{Happy Faces}
\begin{figure}[tb]
\centering
\includegraphics[width=\linewidth]{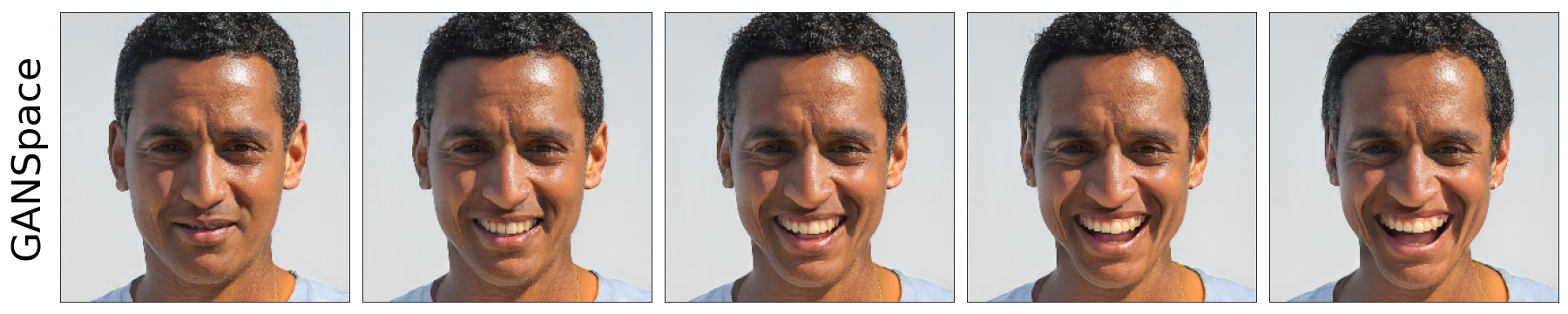}
\includegraphics[width=\linewidth]{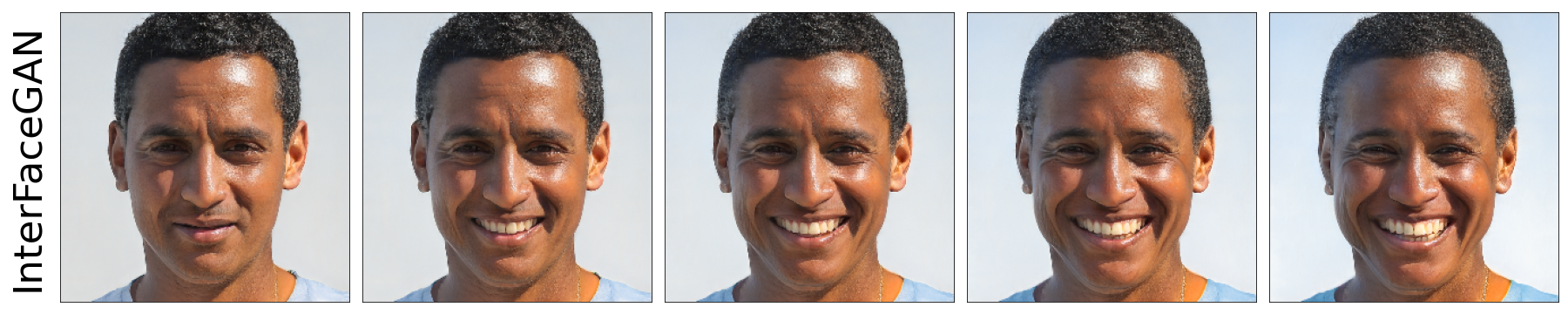}
\includegraphics[width=\linewidth]{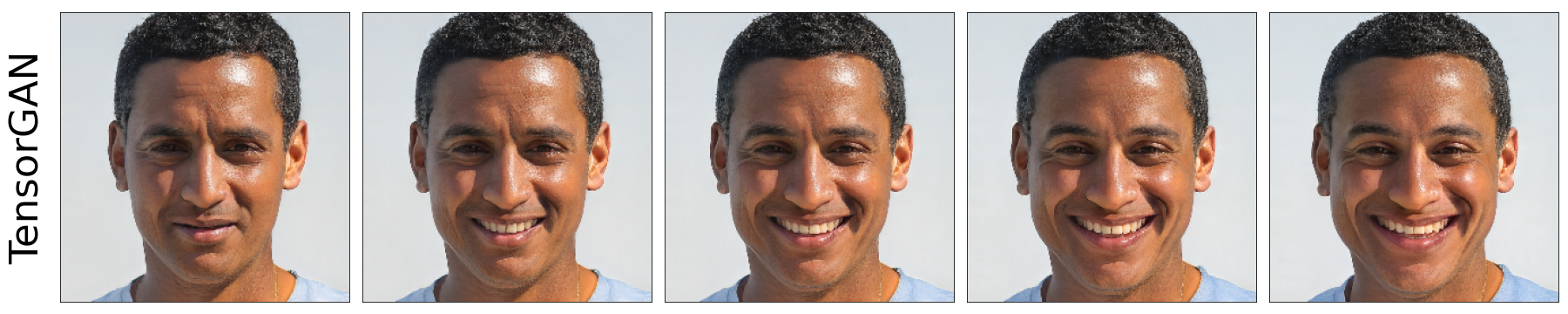}
\caption{Visual comparison of editing a randomly sampled latent code in the smiling directions found in GANSpace \cite{Harkonen2020GANSpace} and InterFaceGAN \cite{Shen2020InterfaceganTPAMI} with the happiness direction found in this work.}
\label{smile_comparrison}
\end{figure}

We compared the found happiness direction to the smile directions from GANSpace and InterFaceGAN, respectively. 
For GANSpace we used the 47$^\text{th}$ principal component applied to the 5$^\text{th}$ and 6$^\text{th}$ style vectors. 
The results are shown in Fig.~\ref{smile_comparrison}. 
Although each method resulted in a smile in the generated image, the style of smile is different. Our method yielded a wider smile whereas GANSpace yielded a smile with a larger mouth opening, while the smile by InterFaceGAN seems to fall between these two.

\subsection{Face Frontalization}
\begin{figure}[tb]
   \centering
   \includegraphics[width=\linewidth]{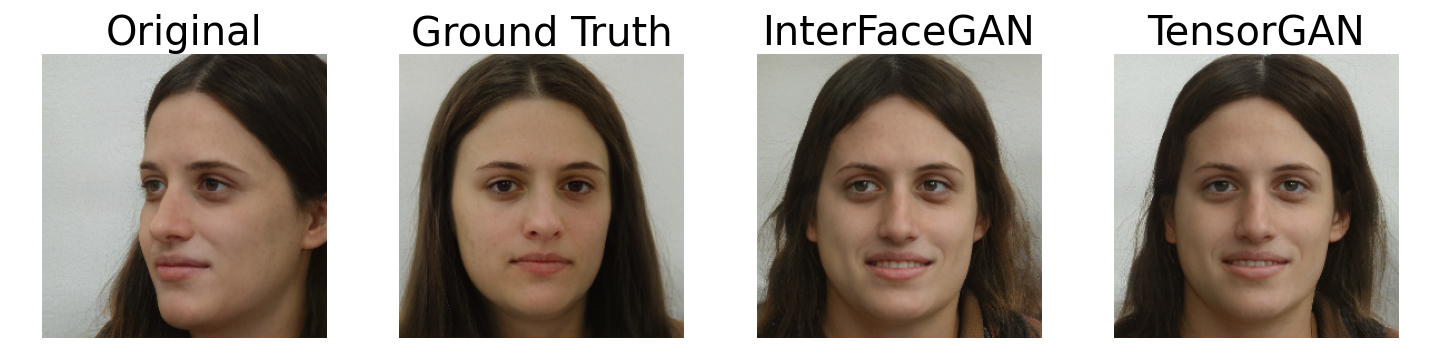}
   \includegraphics[width=\linewidth]{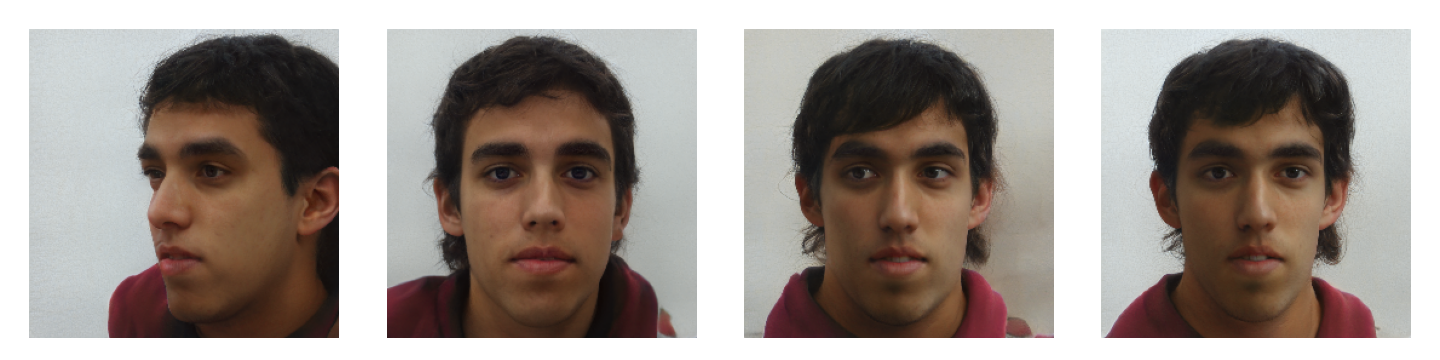}
   \caption{Qualitative comparison of facial frontalization with InterFaceGAN \cite{Shen2020InterfaceganTPAMI} and our method on  
   FEI face database \cite{Ferrari2017Dictionary3DMM}.}
   \label{facefrontalization}
\end{figure}

To experiment face frontalization, we started with the latent codes corresponding to the rotated images in the FEI database \cite{Thomaz2010Feidatabase}, then edited the yaw of latent code to frontalize the images.
Quantative comparison is shown in Fig.~\ref{facefrontalization}. In Tab.~\ref{facefrontalization-stat}, we compare the perceptual and identity similarity scores of the frontalized images to the ground truth. It can be seen the frontalized images 
are very similar to the result obtained by using the pose direction from InterFaceGAN. 
However, our method yielded better similarity scores against to the ground truth.
In addition, the gaze direction by InterFaceGAN is not straight ahead whereas ours is. 

\begin{table}[tb]
\centering
\caption{Comparison of perceptual and identity similarity scores of facial frontalization of images from the FEI face database with InterFaceGAN \cite{Shen2020InterfaceganTPAMI} and our method. The results are reported as mean value $\pm$ standard error of the mean.
}
\label{facefrontalization-stat}
\begin{tabular}{||c c c||} 
\hline
& LPIPS \cite{Zhang2018LPIPS} & ArcFace \cite{Deng2019ArcFace}  \\ [0.5ex] 
\hline\hline
InterFaceGAN & $0.315\pm0.003$ & $0.402 \pm 0.008 $  \\ 
\hline
TensorGAN & $0.305\pm0.004$ &  $0.372 \pm 0.008$ \\
\hline
\end{tabular}
\end{table}

\subsection{Validation with expression classifier}
To validate that the semantic directions recovered with our approach produce a change in the generated images corresponding to the intended labels, we use a pre-trained expression classifier \cite{pyfeat} which is trained on the FER2013 data set \cite{Goodfellow2013Challenges}. 
We sampled $5\times10^3$ random images with varying expressions from StyleGAN and edited these in the direction of each basic emotion. 
Using the classifier, we obtained the probability mass distribution of expressions for the sampled and edited images. 
From this, we calculated the average difference in probability mass due to the edit and visualize the results with a heatmap in Fig.~\ref{recovering_parameters_overview}.

\begin{figure}[t]
\centering
\includegraphics[width=0.78\linewidth]{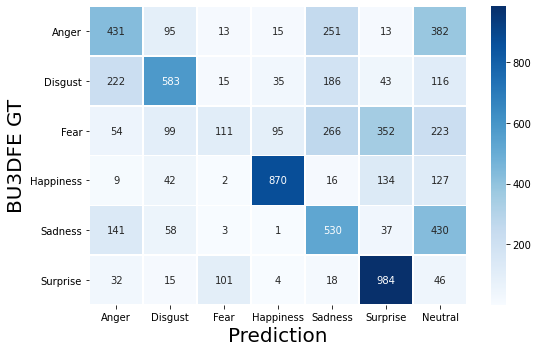}
\caption{Confusion matrix showing the Pyfeat classification results on BU-3DFE. 
It shows that the correlation between Fear/Surprise and Disgust/Anger is not due to a limitation of our model, but can attributed to the differences between the BU-3DFE and FER2013 data sets.
}
\label{fig:bu3dfe_confuss}
\end{figure}

The edits in the direction of anger, happiness, sadness, and surprise lead to changes in the class probabilities which corresponds to an increase in probability of the expected class labels. 
However, the edits in the disgust direction lead to an increase in probability for anger as well as disgust while edits in the fear direction 
leads to a larger probability mass for the surprise label.
This is explained by the fact that PyFeat also classifies the BU-3DFE raw images in a similar way as can be seen in the confusion matrix in Fig.~\ref{fig:bu3dfe_confuss}. Thus, this discrepancy is not due to a limitation of our model, but rather due to systematic differences between the BU-3DFE and FER2013 data sets, which are especially apparent for data points annotated with the fear or disgust labels.

\section{Conclusion}
In this work, we have presented an extension of the HOSVD-based tensor model, proposed in \cite{Haas2021tensorGAN}.
In contrast to \cite{Haas2021tensorGAN}, (1) we use the e4e encoder \cite{Tov2021e4e} to recover highly editable latent codes for the BU-3DFE database, (2) we improve reconstruction in the tensor model by allowing the parameters to be full-rank, and (3) we show that edits can be applied directly in latent space.
Further, we showed that we can calculate linear directions in  latent space corresponding to the six prototypical emotions by truncating the emotion intensity subspace.
%
After obtaining a latent representation of the data, constructing the tensor model is fast, requiring only a few minutes to calculate the HOSVD.
Further, the latent space directions corresponding to the six prototypical emotions can be calculated from the tensor model and subsequently applied to any latent code in the original latent space without the need to first estimate the subspace parameters  as otherwise suggested in \cite{Haas2021tensorGAN}. In other words, the found semantic directions are global and can be applied to any latent code without any further calculations. Our
%
%
\begin{figure}[t]
\centering
\includegraphics[width=0.78\linewidth]{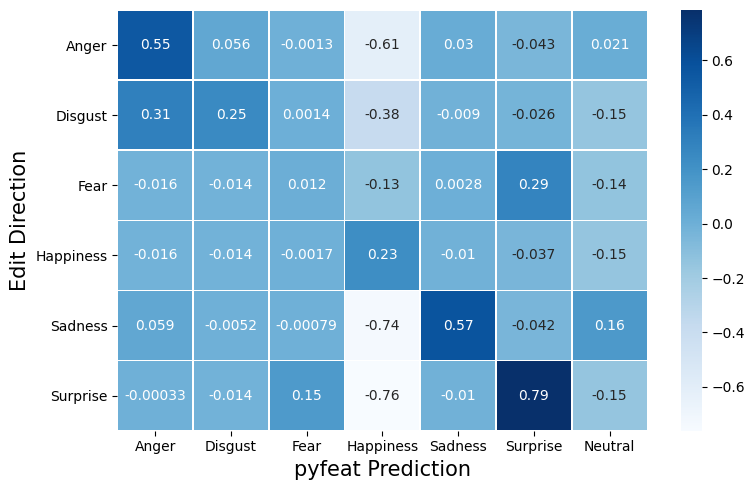}
\caption{Heatmap of the average difference in expression probability masses due to expression edits with our approach. 
Note that Fear increases the probability mass for Surprise and Disgust increases the probability mass for Anger. The reason is explained in the main text.
} \vspace{-2mm}
\label{recovering_parameters_overview}
\end{figure}
%
%
\hspace{-2mm} method is able to identify directions in latent space corresponding to yaw rotation, as well as each of the six basic expressions. 
The quality of the edits performed with these directions is on par with the corresponding edits using GANSpace \cite{Harkonen2020GANSpace} and InterFaceGAN \cite{Shen2020InterfaceganTPAMI}. 
